# A CONSTRAINT-BASED CASE FRAME LEXICON ARCHITECTURE

Kemal Oflazer and Okan Yılmaz
Department of Computer Engineering and Information Science
Bilkent University
Bilkent, Ankara 06533, Turkey
{ko,okan}@cs.bilkent.edu.tr
Fax:(90-312) 266-4126

## 1  Introduction

Recent advances in theoretical and implementational aspects of feature and constraint-based formalisms for representing linguistic information have fostered research on the use of such formalisms in the design and implementation of computational lexicons [1]. Case-frame approach has been the representation of choice especially for languages with free constituent order, explicit case marking of noun phrases and embedded clauses filling nominal syntactic roles. The semantics of such syntactic role fillers are usually determined by their lexical semantic and morphosyntactic properties, instead of position in the sentence [5]. In this paper we present our approach to building a constraint-based case frame lexicon for use in natural language processing in Turkish.

A number of observations that we have made on Turkish have indicated that we have to go beyond the traditional transitive and intransitive distinction, and utilize a framework where verb valence is considered as the obligatory co-existence of an arbitrary subset of possible arguments along with the obligatory exclusion of certain others, relative to a verb sense. Additional morphosyntactic, lexical and semantic constraints are utilized to map a given syntactic structure to a specific verb sense.

In the next sections, we will first present some motivating observations from Turkish. We will then present the architecture of the case frame lexicon and then continue by describing the constraint structure. After giving some examples, we will present our conclusions and suggestions.

## 2  Issues in Representing Case-Frame Information

In Turkish, (and possibly in many other languages) verbs often convey several meanings (some totally unrelated) when they are used with subjects, objects, oblique objects, adverbial adjuncts, with certain lexical, morphological, and semantic features, and co-occurrence restrictions. In addition to the usual sense variations due to selectional restrictions on verbal arguments, in most cases, the meaning conveyed by a case frame is idiomatic and not compositional, with subtle constraints. For example, the Turkish verb *ye* (*eat*), when used with a direct object noun phrase whose head is:

1. *para* (*money*), with *no* case or possessive markings and a human subject, means *to accept bribe*,

2. *para* (*money*), with a non-human subject, means *to cost a lot*,

3. *para* (or any other NP whose head is ontologically IS-A money, e.g., *dolar, mark*, etc.) with *obligatory* accusative marking and *optional* possessive marking, means *to spend money*,

4. *kafa* (*head*) with *obligatory* accusative marking and *no* possessive marking, means *to get mentally deranged*,

5. *hak* (*right*) with optional accusative and possessive markings, means *to be unfair*,

6. *baş* (*head*, cf. 4) (or any NP whose head is ontologically IS-A human) with *optional* accusative and optional possessive marking (obligatory only with *baş*), means *to waste* or *demote a person*.

Clearly such usage has impact on thematic role assignments to various role fillers, and even on the syntactic behavior of the verb in question. For instance, for the third and fourth cases where the object has to be obligatorily case-marked accusative, a passive form would not be grammatical for the sense conveyed although syntactically *ye* is a transitive verb. If, furthermore, other objects are also present then sense resolution becomes a bit more subtle. Again for the same verb *ye*:

1. if an *ablative case-marked oblique object* denoting an edible entity is present, then there should *not* be any direct object, and the verb sense maps to *to eat a piece of (the edible (oblique) object)*.

2. if the ablative case-marked oblique object does not denote something edible, but rather a container, then the sense maps to *to eat out of*, with the *optional* direct (edible) object denoting the object eaten.

Sometimes verbs require different combinations of arguments, or explicitly require that certain arguments not be present. For instance, the verb *şaş* chooses different arguments depending on the sense, obligatorily excluding other objects. For instance when *şaş* is used with

1. an *ablative case-marked oblique object* and with no other object in the case frame, it means *to deviate from*,

2. a *dative case-marked oblique object* and with no other object, it means *to be surprised at*,

3. an *accusative case-marked* direct object with no other object, it means *to be confused about*.

Another example along the same lines but with a different flavor is for the verb *geç* (to pass).

1. When used with a *causative marking on the verb* and with an *accusative case-marked direct object* (of the causative verb) and *no ablative case-marked oblique object*, the meaning maps to *to see someone off*.

2. If, however the *ablative case-marked object* is present then the sense resorts to the main sense of the verb *to pass something from someplace (to someplace)* (this sense being a much weaker alternative interpretation of the former case.)

As illustrated in the examples above, verb sense resolution and idiomatic usage determination has to be dealt with in a principled way. In this paper, we present a unification-based approach to a constraint-based case frame lexicon, for use in natural language processing in Turkish. The essential function of our lexicon is to map a case frame containing information that is essentially syntactic, to a semantic frame which captures the predication denoted by the case frame along with information about who fills what thematic role in that predication. This is an extension of our previous work [9, 10], and is to be used in our efforts on parsing and machine translation[2, 6].

## 3 The Lexicon Architecture

In this section we present an overview of structure of lexicon entries and the nature of the constraints that are used to resolve case frame meanings. The basic unit in the lexicon is a *sense* which is the information denoting an indivisible predication along with the thematic roles involved. We generate the case frame of each sense by unifying a set of co-occurrence, morphological, syntactic, semantic, and lexical constraints. The lexicon is implemented in TFS [3] by the disjunction of the senses defined by unifying `wf-case-frame` (well-formed case frame) with each sense:

```
        wf-case-frame   <   case-frame.

    wf-case-frame    &    SENSE#1.
    wf-case-frame    &    SENSE#2.
        ...                ...
    wf-case-frame    &    SENSE#n.
```

By defining case-frame as the disjunction of all senses, we in fact build an active lexicon. Note that this lexicon can be used bidirectionally for mapping from a syntactic case frame to a verb sense and thematic role filler, and vice versa. The latter may be used in syntactic generation when a verb sense has been mapped to one of the entries via a transfer process.

## 3.1 Lexicon Entries

Each entry in our lexicon has the structure shown by the attribute-value matrix in Figure 1.

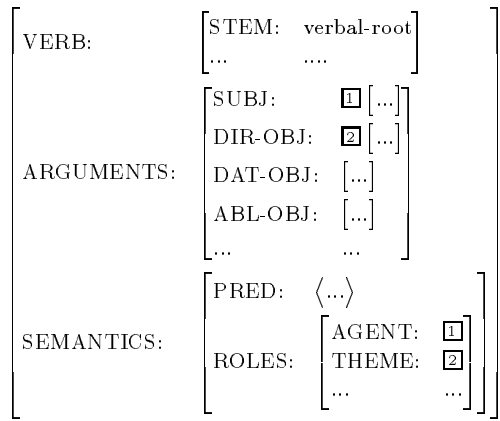

Figure 1: Structure of a case frame lexicon entry.

The feature structure for each argument contains information about the morphological and syntactic structure about the syntactic constituents such as major and minor part-of-speech category, agreement, case, possessive markers, additional morphological markings such as verb form, (e.g., infinitive, participle, etc.), voice (e.g. active, passive, causative, reflexive, reciprocal, etc.) for embedded S's, along with their own case frames.[1] This structure is similar to the structure proposed in Lascarides et. al. [4]. However, instead of classifying argument structures as simply transitive, intransitive, etc., we consider all relevant elements of the power set of all possible arguments. For Turkish, the arguments that we have chosen to include in the argument slot (for a verb in active voice) comprises the following:

- subject (nominative NP),

- direct object (nominative or accusative case-marked NP denoting theme or patient)

- oblique objects (ablative, dative, locative case-marked NP denoting source, goal, location, unless overridden by a specific sense)

- beneficiary object (dative case-marked NP, or PP with a certain PFORM)

---

[1] For instance, if the verb is *tut* (catch/hold) with obligatory $3^{rd}$ person singular agreement and active voice, and the subject is a (nominalized) S with a verb form future participle, then the sense conveyed by the top level case frame is *to feel like doing* the predication indicated by the subject S's case frame, with the agent being the subject of this embedded clause.

- instrument object (instrumental case-marked NP or PP with a certain PFORM)
- value object (dative case-marked NP or PP with a certain PFROM)

For example, in (1) *götür* (*to take from some place to some place*) is used with all arguments it subcategorizes for. Here *ben* (*I*) is the subject, *çocuğu* (*the child*) is the direct object, *evden* (*from home*) and *okula* (*to school*) are oblique objects, *annesi için* (*for his/her mother*) is the beneficiary, *otobüs* (*bus*) is the instrument, and *10 lira* (*10 liras*) is the value-designator[2] of *götür*.

(1)

a. Ben çocuğu annesi için evden okula otobüsle liraya götürdüm.

b. I child+ACC mother+3SG for house+ABL school+DAT bus+INS 10 lira+DAT take+PAST+1SG

c. I took the child from home to school by bus for 10 liras for his/her mother.

In general, there may be more than one instantiation of the SEMANTICS frame for a given instantiated set of case frame arguments, but usually there will be a preference ordering among the solutions for a given case frame. For instance, for the *ye* verb discussed above, the argument structure for the third case giving rise to the meaning *to get mentally deranged* may conceivably give rise to a literal meaning in a rather improbable context (such as eating the head of a fish at dinner - much in the spirit of the two interpretations of the English idiom *kick the bucket*.)

## 3.2 Constraint-Architecture

We express constraints on the arguments in the case frame of a verb via a 5-tier constraint hierarchy:

1. *Constraints on verb features*: These describe any relevant constraints on the morphological features of the verb. For instance, for the example in footnote 1, the verb has to have the $3^{rd}$ person singular agreement for that sense.

2. *Constraints on morphological features*: These describe any obligatory constraints on the arguments, such as case-marking, verb form (in the case of embedded clauses), etc.

3. *Constraints on argument co-occurrence*: In general verb senses and idiomatic usage are determined by constraints that indicate which arguments *should, can* and *not* occur together in order to convey a specific meaning. At this level, one expresses *obligatory* argument co-occurrence constraints along with constraints that indicate when certain arguments should *not* occur.

   Any arguments that are not obligatorily required by for resolving a specific verb sense, hence optional, serve to modify the sense in a number of aspects. For instance an instrumental object is not obligatory for resolving a case frame of *ye* to *to eat*. Existence of such an object would modify the basic semantics to *to eat with (instrument)*, provided selectional restriction constraints are met.

4. *Lexical constraints*: These indicate any specific constraints on the heads of the arguments in order to convey a certain sense, and usually constrain the stem of the head noun to be a certain lexical form, or one of a small set of lexical forms.

5. *Semantic Constraints*: These indicate essentially selectional restriction constraints that may in general be resolved using a companion ontological database. In this ontological database, we model the world by defining semantic categories, such as *human, thing, non-living object, living object*, etc. An interconnected network of these semantic categories is built in TFS by using the multiple inheritance mechanism provided. A simple lexicon keeping the semantic markers of some commonly used words is defined. The ontological database is based on the ontology in Nagao, Tsujii and Nakamura [5]. Semantic markers for words are defined under ten major concepts:

- `Thing-Object` containing matters such as things and objects,
- `Commodity-Ware` containing artificial matters useful to humans,
- `Idea-Abstraction` containing non-matters which results from intellectual activities in the human brain, such as ideas and abstractions,
- `Part` containing structural parts, elements, and contents of things and matters,
- `Attribute` containing properties, qualities, or features which are representatives of things,
- `Phenomenon` containing physical, chemical, and social actions,
- `Doing-Action` containing human actions,
- `Sentiment-MentalActivity` containing humans' mental activities,
- `Measure` containing measures, and
- `Time-Space` containing time, space, and topologies.

---
[2]Almost all the Turkish verbs can be accompanied by a *value-designator*.

Each of these groups consist of subconcepts (e.g., `Human-Role-Profession` (human, role, or profession), `Animal`, `Emotion`, etc.).

We represent the constraints as a constraint hierarchy sharing constraints among the specification of other constraints and sense definitions, whenever possible. Figure 2 illustrates the simplified form of the constraint-sense hierarchy of the verb *ye* (*eat*). Each sense was defined as unification of co-occurrence, morphological, syntactic, semantic, and lexical constraints. In this figure, the arrows are from the constraint frames to senses inheriting them.

## 3.3 Examples

Here we present a very simple example that shows how one can describe a given verb meaning:
The following constraints are employed:

1. `VERB-IS-YE` is a verb constraint corresponding to the feature structure

    [VERB: | STEM: "ye"]

2. `DIR-OBJ-HAS-NO-POSS` is the morphological constraint equivalent to the feature structure

    [ARGUMENTS: | DIR-OBJ: | POSS: none]

3. `DIR-OBJ-IS-ACC` is a morphological constraint equivalent to the feature structure

    [ARGUMENTS: |DIR-OBJ: |CASE: acc]

4. `NO-DATIVE-OBL-OBJ` is an argument co-occurrence constraint equivalent to the feature structure

    [ARGUMENTS: |DAT-OBL: nil]

5. `SUBJECT-IS-HUMAN` is a semantic constraint equivalent to

    [ARGUMENTS: |SUBJECT: |HEAD: |SEM: human]

6. `DIR-OBJ-HEAD-LEX-KAFA` is a lexical constraint

    [ARGUMENTS: |DIR-OBJ: |HEAD: |LEX: "kafa"]

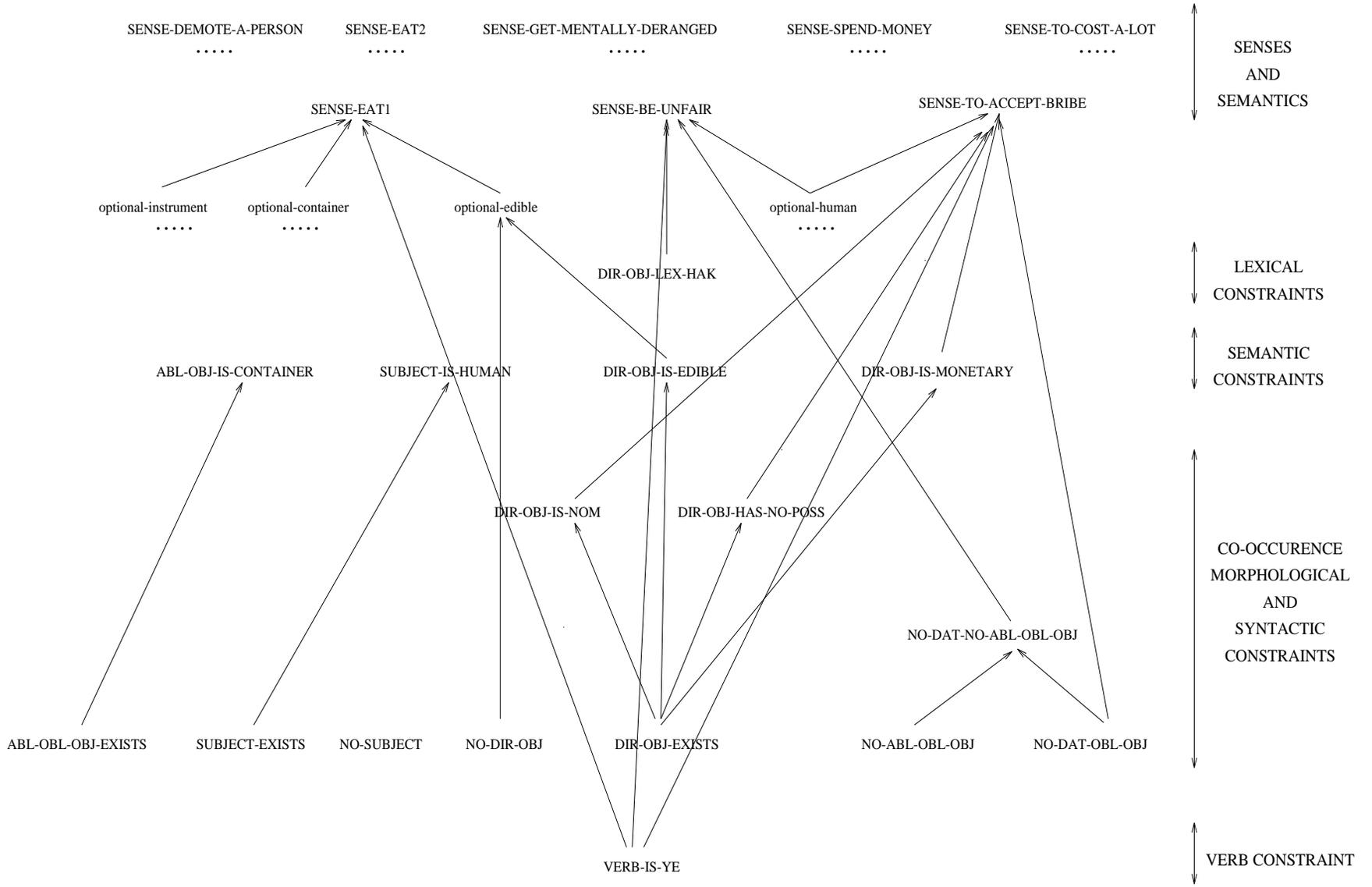

Figure 2: Inheritance mechanism among senses and the constraints of the Turkish verb *"ye"*.

7. `SEM-GET-MENTALLY-DERANGED` is the feature structure for the semantics portion

$$\begin{bmatrix} \text{ARGUMENTS:} & [\text{SUBJ:} \quad \boxed{1}] \\ \text{SEMANTICS:} & \begin{bmatrix} \text{PRED:} & \text{``get mentally deranged''} \\ \text{ROLES:} & [\text{EXPERIENCER:} \quad \boxed{1}] \end{bmatrix} \end{bmatrix}$$

We can then express the constraint for a given verb sense by unifying (denoted by & in TFS) all the constraints above:

```
SENSE-GET-MENTALLY-DERANGED  :=
            VERB-IS-YE              &   DIR-OBJ-HAS-NO-POSS   &
            DIR-OBJ-IS-ACC          &   NO-DATIVE-OBL-OBJ     &
            DIR-OBJ-LEX-KAFA        &   SUBJECT-IS-HUMAN      &
            SEM-GET-MENTALLY-DERANGED
```

The resulting constraint when unified with partially specified case frame entry – an entry where only the argument and verb entries have been specified, will supply the unspecified SEMANTICS component(s). That is, when a partially specified case frame such as

$$\begin{bmatrix} \text{VERB:} & [\text{STEM:} \quad \text{``ye''}] \\ \text{ARGUMENTS:} & \begin{bmatrix} \text{SUBJ:} & \begin{bmatrix} \text{CAT:} & \text{NP} \\ \text{HEAD:} & \begin{bmatrix} \text{CAT:} & \text{N} \\ \text{STEM:} & \text{``adam''} \\ \text{CASE:} & \text{nom} \\ \text{AGR:} & \text{3SG} \\ \text{POSS:} & \text{none} \end{bmatrix} \\ \text{MOD:} & \text{nil} \end{bmatrix} \\ \text{DIR-OBJ:} & \begin{bmatrix} \text{CAT:} & \text{NP} \\ \text{HEAD:} & \begin{bmatrix} \text{CAT:} & \text{N} \\ \text{STEM:} & \text{``kafa''} \\ \text{CASE:} & \text{acc} \\ \text{AGR:} & \text{3SG} \\ \text{POSS:} & \text{none} \end{bmatrix} \\ \text{MOD:} & \text{nil} \end{bmatrix} \end{bmatrix} \end{bmatrix}$$

unifies successfully with the given constraint above, will instantiate the unspecified portion to

$$\begin{bmatrix} \ldots \\ \text{SEMANTICS:} & \begin{bmatrix} \text{PRED:} & \text{``get mentally deranged''} \\ \text{ROLES:} & \begin{bmatrix} \text{EXPERIENCER:} & \begin{bmatrix} \text{STEM:} & \text{``adam''} \\ \text{CAT:} & \text{N} \\ \text{CASE:} & \text{nom} \\ \text{AGR:} & \text{3SG} \\ \text{POSS:} & \text{none} \end{bmatrix} \end{bmatrix} \end{bmatrix} \end{bmatrix}$$

As a second example, consider:

(2)

   a. Demet pasta yedi.

   b. Demet cake eat+3SG+PAST

   c. Demet ate cake.

where *ye* is used in sense *to eat*. The input and output case-frames for the sentence (2) are given in Figures 3 and 4, respectively. In this example, the constraints are:

1. `VERB-IS-YE` is a verb constraint corresponding to the feature structure

   [VERB: | STEM: "ye"]

2. `NO-DAT-OBL-OBJ` is a co-occurrence constraint equivalent the feature structure:

   [ARGUMENTS: |DAT-OBL: nil]

3. `DIR-OBJ-IS(optional-edible)` is a constraint equivalent to

   [ARGUMENTS: |DIR-OBJ: |HEAD: |SEM: edible] or [ARGUMENTS: |DIR-OBJ: nil]

   (This is just explanatory, see below for how this is implemented in TFS.)

4. `ABL-OBJ-IS(optional-container)` is a constraint equivalent to

   [ARGUMENTS: |ABL-OBJ: |HEAD: |SEM: container] or [ARGUMENTS: |ABL-OBJ: nil]

   (This is again handled similar to above).

5. `INST-OBJ-IS(optional-instrument)` is a constraint equivalent to

   [ARGUMENTS: |INST: |HEAD: |SEM: instrument] or [ARGUMENTS: |INST: nil]

   (This is again handled similar to above).

6. `SEM-EAT` is the feature structure for the semantics portion

$$\begin{bmatrix} \text{ARGUMENTS:} & \begin{bmatrix} \text{SUBJ:} & \boxed{1} \\ \text{DIR-OBJ:} & \boxed{2} \\ \text{ABL-OBJ:} & \boxed{3} \\ \text{INST:} & \boxed{4} \end{bmatrix} \\ \text{SEMANTICS:} & \begin{bmatrix} \text{PRED:} & \text{"to eat"} \\ \text{ROLES:} & \begin{bmatrix} \text{AGENT:} & \boxed{1} \\ \text{THEME:} & \boxed{2} \\ \text{SOURCE:} & \boxed{3} \\ \text{INSTRUMENT:} & \boxed{4} \end{bmatrix} \\ \text{EXAMPLE:} & \text{"biz (çatalla) (dolaptan) (pasta) yedik."} \end{bmatrix} \end{bmatrix}$$

When the partially specified case-frame in Figure 3 is unified with the constraints stated above, the case-frame shown in Figure 4 is obtained. In most of the cases there exist optional arguments, that is, arguments that are not obligatorily required for resolving a verb sense. These, nevertheless, have to be constrained, usually on semantic grounds. For instance the direct object is not obligatory for the basic sense of *ye*, but has to be an edible entity if it is present. We handle these constraints by defining a slightly more complex type hierarchy:

```
argument = noun-phrase | case-frame | optional.
optional = optional-edible | optional-container | optional-instrument :..
optional-edible = nil | edible-obj.
edible-obj & noun-phrase & IS-A-EDIBLE.
```

where `IS-A-EDIBLE` is a constraint on the `noun-phrase` type of the form [HEAD: | SEM: edible].
The optional ablative and instrumental objects are defined similarly. The sense definition becomes:

```
SENSE-EAT1 := VERB-IS-YE & NO-DATIVE-OBLIQUE-OBJ & DIR-OBJ-IS(optional-edible)
& ABL-OBL-OBJ(optional-container)& INST-OBJ-IS(optional-instrument) & SEM-EAT1.
```

As a more complicated example employing nested clauses, we present in Figure 5, the case frame for the example in footnote 1, where the verb *tut* (*catch*) is used with a clausal subject for a very specific sense. In this case, the sense resolution of the embedded case-frame is also performed concurrently with the case frame resolution of the top-level frame.

$$\begin{bmatrix} \text{VERB:} & \begin{bmatrix} \text{CAT:} & \text{V} \\ \text{STEM:} & \text{"ye"} \end{bmatrix} \\ \text{ARGUMENTS:} & \begin{bmatrix} \text{SUBJ:} & \begin{bmatrix} \text{CAT:} & \text{NP} \\ \text{HEAD:} & \begin{bmatrix} \text{CAT:} & \text{N} \\ \text{STEM:} & \text{"demet"} \\ \text{CASE:} & \text{nom} \\ \text{AGR:} & \text{3sg} \end{bmatrix} \\ \text{MOD:} & \text{nil} \end{bmatrix} \\ \text{DIR-OBJ:} & \begin{bmatrix} \text{CAT:} & \text{NP} \\ \text{HEAD:} & \begin{bmatrix} \text{CAT:} & \text{N} \\ \text{STEM:} & \text{"pasta"} \\ \text{CASE:} & \text{nom} \\ \text{POSS:} & \text{none} \end{bmatrix} \\ \text{MOD: nil} \end{bmatrix} \\ \text{ABL-OBJ:} & \text{nil} \\ \text{DAT-OBJ:} & \text{nil} \\ \text{INST:} & \text{nil} \end{bmatrix} \end{bmatrix}$$

Figure 3: The input syntactic frame for the sentence (2)

$$\begin{bmatrix} \text{VERB:} & \begin{bmatrix} \text{CAT:} & \text{V} \\ \text{STEM:} & \text{"ye"} \end{bmatrix} \\ \text{ARGUMENTS:} & \begin{bmatrix} \text{SUBJ:} & \boxed{1}= \begin{bmatrix} \text{CAT:} & \text{NP} \\ \text{HEAD:} & \begin{bmatrix} \text{CAT:} & \text{N} \\ \text{STEM:} & \text{"demet"} \\ \text{CASE:} & \text{nom} \\ \text{AGR:} & \text{3sg} \\ \text{SEM:} & \text{female} \end{bmatrix} \\ \text{MOD:} & \text{nil} \end{bmatrix} \\ \text{DIR-OBJ:} & \boxed{2}= \begin{bmatrix} \text{CAT:} & \text{NP} \\ \text{HEAD:} & \begin{bmatrix} \text{CAT:} & \text{N} \\ \text{STEM:} & \text{"pasta"} \\ \text{CASE:} & \text{acc} \\ \text{POSS:} & \text{none} \\ \text{SEM:} & \text{edible} \end{bmatrix} \\ \text{MOD:} & \text{nil} \end{bmatrix} \\ \text{OBL-ABL:} & \boxed{3}= \text{nil} \\ \text{INST:} & \boxed{4}= \text{nil} \\ \text{OBL-DAT:} & \text{nil} \end{bmatrix} \\ \text{SEMANTICS:} & \begin{bmatrix} \text{PRED:} & \text{"to eat"} \\ \text{ROLES:} & \begin{bmatrix} \text{AGENT:} & \boxed{1} \\ \text{THEME:} & \boxed{2} \\ \text{SOURCE:} & \boxed{3} \\ \text{INSTRUMENT:} & \boxed{4} \end{bmatrix} \\ \text{EXAMPLE:} & \text{"biz (çatalla) (dolaptan) (pasta) yedik."} \end{bmatrix} \end{bmatrix}$$

Figure 4: The output syntactic frame for the sentence (2)

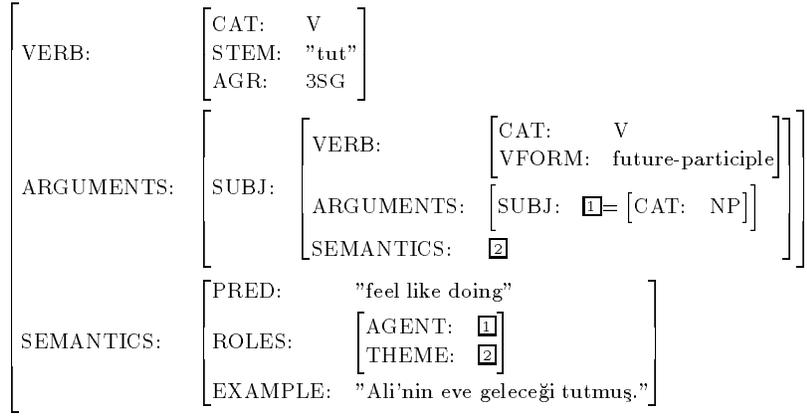

Figure 5: Fully instantiated case frame for the example in footnote 1.

## 4 Related Work

In recent years, there have been several studies on constraint-based lexicons. Russell et. al. proposes an approach to multiple default inheritance for unification-based lexicon [8]. In this study, the ECL lexicon is developed by using the advantages of both default inheritance and unification. The default inheritance is achieved by nested generalizations with exceptions. When multiple inheritance is allowed the order becomes important, although unification is an unordered mechanism. This system resolves the interaction problems such as cycles, ambiguity, and the redundancy of multiple paths by defining a total ordering on superclasses. In our system, we deal with this problem by unifying sub-frames on distinct constraints. In another study by Lascarides et. al. [4], an ordered approach to default unification is suggested. Our system is similar to theirs. However, in our system because of the characteristics of argument structures of verbs in Turkish, we did not define default types as specific as used there. Finally, De Paiva [7] formalizes the system of well-formed typed feature structures. In this study, type hierarchies and relations are mathematically defined. They also formalize unification and generalization operators between the feature structures, along with defining well-formedness notion that we used in our system.

## 5 Conclusions

This paper has presented an overview of our research for representing verb sense resolution using constraints on syntactic case frames. We express constraints leading to a given sense by a series of constraints on different dimensions of the information available, and achieve economy of representation via sharing of constraints across many verb sense definitions. Although we have not yet formulated it, incorporation of constraints on adjuncts can also be done with relative ease. This would for example distinguish among:

1. *Bu resim buraya düştü*. (This picture fell here.), vs.

2. *Bu resim buraya* **iyi** *düştü*. (This picture looks good here.), vs.

3. *Bu resmi* **iyi** *düş***ür***düm*. (I got this picture cheaply.)

where the idiomatic uses are triggered by the adverbial *iyi* (good), or its combination with the causative voice marker (-ür-) on the verb.

Another detail that we have not covered here is how various verb valence changing transformations such as passivizations and causativizations are dealt with. We are currently working on this aspect of the problem and on further theoretical investigation of this mapping from the space of syntactic structures (that of the power set of possible arguments) to semantic senses, and on the mechanisms for ordering the multiple semantic component instantiations for a given case-frame.

# 6  Acknowledgments

This research was supported in part by a NATO Science for Stability Project Grant, TU-LANGUAGE.

# References

[1] E.J. Briscoe, A. Copestake, and Ve. de Paiva (eds.). 1993. *Inheritance Defaults and the Lexicon*. Cambridge University Press.

[2] Z. Güngördü and K. Oflazer. Parsing Turkish using the lexical-functional grammar formalism. In *Proceedings of COLING-94, the 15th International Conference on Computational Linguistics*, Kyoto, Japan, 1994.

[3] J. Kuhn. *Encoding HPSG Grammars in TFS*. Institut für Maschinelle Sprachverarbeitung, Universität Stuttgart, Germany, March 1993.

[4] A. Lascarides, T. Briscoe, N. Asher, and A. Copestake. Order Independent and Persistent Typed Default Unification, Technical Report, Cambridge University, Computer Laboratory, March 1995.

[5] M. Nagao, J. Tsujii, and J. Nakamura. The Japanese Government Project for Machine Translation. In *Computational Linguistics*, volume 11. April-September 1985.

[6] K. Oflazer and C. Bozşahin. Turkish Natural Language Processing Initiative: An Overview. In *Proceedings of the Third Turkish Symposium on Artificial Intelligence and Neural Networks*, June 1994.

[7] V. de Paiva. Types and Constraints in LKB. In E.J. Briscoe, A. Copestake, and Ve. de Paiva (eds.), *Inheritance Defaults and the Lexicon*, Cambridge University Press, Cambridge, England, pp. 164-189.

[8] G. Russell, A. Ballim, J. Carroll, and S. Warwick-Armstrong. A Practical Approach to Multiple Default Inheritance for Unification-Based Lexicons. In E.J. Briscoe, A. Copestake, and Ve. de Paiva (eds.), *Inheritance Defaults and the Lexicon*, Cambridge University Press, Cambridge, England, pp. 137-147.

[9] O. Yılmaz. Design and implementation of a verb lexicon and a verb sense disambiguator for Turkish. Master's thesis, Department of Computer Engineering and Information Sciences, Bilkent University, Ankara, Turkey, September 1994.

[10] O. Yılmaz and K. Oflazer. Design and implementation of a verb lexicon and a verb sense disambiguator for Turkish. In *Proceedings of the Fourth Turkish Symposium on Artificial Intelligence and Neural Networks*, June 1995.